\documentclass{article}

\usepackage{microtype}
\usepackage{graphicx}
\usepackage{amsmath}

\usepackage{subcaption}

\usepackage{booktabs} 

\usepackage{hyperref}

\usepackage[accepted]{icml2018}

\icmltitlerunning{Data Extraction from Charts via Single Deep Neural Network}

\begin{document}

\twocolumn[
\icmltitle{Data Extraction from Charts via Single Deep Neural Network}



\icmlsetsymbol{equal}{*}

\begin{icmlauthorlist}
\icmlauthor{Xiaoyi Liu}{nu}
\icmlauthor{Diego Klabjan}{nu}
\icmlauthor{Patrick N Bless}{intel}
\end{icmlauthorlist}

\icmlaffiliation{nu}{Department of Industrial Engineering and Management Science, Northwestern University, Evanston, USA}
\icmlaffiliation{intel}{Intel Corporation, Chandler, USA}

\icmlcorrespondingauthor{Xiaoyi Liu}{xiaoyiliu2021@u.northwestern.edu}
\icmlcorrespondingauthor{Diego Klabjan}{d-klabjan@northwestern.edu}
\icmlcorrespondingauthor{Patrick N Bless}{patrick.n.bless@intel.com}

\icmlkeywords{Machine Learning, ICML}

\vskip 0.3in
]



\printAffiliationsAndNotice{}  

\begin{abstract}

Automatic data extraction from charts is challenging for two reasons: there exist many relations among objects in a chart, which is not a common consideration in general computer vision problems; and different types of charts may not be processed by the same model. To address these problems, we propose a framework of a single deep neural network, which consists of object detection, text recognition and object matching modules. The framework handles both bar and pie charts, and it may also be extended to other types of charts by slight revisions and by augmenting the training data. Our model performs successfully on 79.4\% of test simulated bar charts and 88.0\% of test simulated pie charts, while for charts outside of the training domain it degrades for 57.5\% and 62.3\%, respectively.

\end{abstract}

\section{Introduction}
\label{submission}
``Data everywhere information nowhere" is a common saying in the business world. Consider all of the presentations and reports lingering in folders of a company. They are embellished with eye-appealing charts as images forming formidable data, but getting information from these charts is challenging. To this end, a system that automatically extracts information from charts would provide great benefits in knowledge management within the company. Such knowledge can be combined with other data sets to further enhance business value. We address this problem by developing a deep learning model that takes an image of a chart as input and it extracts information in the form of categories being displayed, the relevant text such as the legend, axis labels, and numeric values behind the data displayed.

There are existing tools for information extraction from charts \cite{Revision, Huang2007} based on traditional computer vision methods, which rely on  complicated human-defined rules and thus are not robust. With the proliferation of deep learning, it is conceivable that the accuracy of chart component detection can be improved without complicated rules, i.e., by using raw images as input with no feature engineering or employment of other rules. Despite of this belief there is still lack of a single deep learning model for data extraction from charts.

We introduce a deep learning solution that automatically extracts data from bar and pie charts and essentially converts a chart to a relational data table. The approach first detects the type of a chart (bar or pie), and then employs a single deep learning model that extracts all of the relevant components and data. There is one single model for bar charts and a different one for pie charts. 
The entire framework has three stages: 1. chart type identification, 2. element detection, text recognition, and bounding box matching through which the actual numerical data is extracted, and 3. inference. The first phase is a standard image classification problem. 
The most interesting part is the second phase where we rely on the Faster-RCNN model \cite{Ren2015FasterRCNN}. 
We add several components to the feature maps of regional proposals, e.g., text detection and recognition. 
The most significant part is the addition of relation network components that match, e.g., part of the legend with a matching bar, a bar with the y-axis value, a slice in the pie chart with part of the legend. 
In order to make extraction from pie charts work, additional novel tricks are needed; e.g., the model detects the angle of each slice by attaching an RNN to regional proposals (since slices form a sequence when traversed in a clockwise manner), multiplies the feature map matrix of the regional proposal of the entire pie with an angle-dependent rotation matrix, and then uses this rotated matrix in the relation network. The last inference phase is using heuristics to produce the final objects and data. 

The model is trained on simulated charts based on Microsoft Excel and the Matplotlib library. It is then evaluated on simulated test data, the Microsoft FigureQA charts data set \cite{Figureqa}, and manually inspected charts from Google Images. The results on the simulated test set show our model performs successfully on 79.4\% simulated bar charts and 88.0\% on simulated pie charts. On charts from FigureQA and Google Images the performance drops to 57.5\% and 62.3\% for bar and pie charts, respectively.

Our main contributions are as follows. 
\begin{itemize}
    \item We propose a single deep learning model that extracts information from bar charts. The new ideas of the model are the combination of text recognition, text detection, and pairwise matching of components within a single model. In particular, we design a new approach for matching candidate components, e.g, an actual bar bounding box with an entry in the legend.
    \item We also propose another single deep learning model for pie charts. This model introduces an RNN component to detect angles in a pie chart, and a different strategy for matching non-rectangular patches. 
    \item We use a pipeline where we first identify a chart type by standard CNN-based classification. Once the chart type is identified, we employ one of the aforementioned models to extract information. 
\end{itemize}

\begin{figure}
\centering
  \begin{subfigure}{0.22\textwidth}
    \centering
    \includegraphics[width=\textwidth]{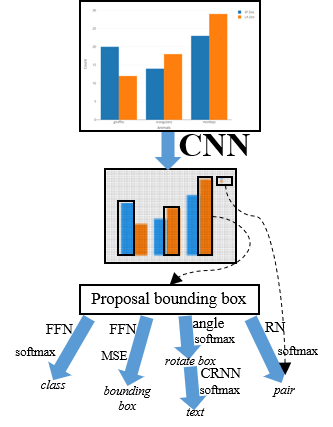}
    \caption{Bar chart extraction}
    \label{fig:1a}
  \end{subfigure}
  \begin{subfigure}{0.245\textwidth}
    \centering
    \includegraphics[width=\textwidth]{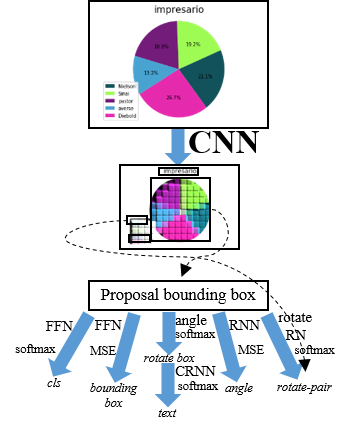}
    \caption{Pie chart extraction}
    \label{fig:1b}
  \end{subfigure}
  \caption{Framework for charts data extraction} 
  \label{fig1}
\end{figure}

In Section~\ref{literature review}, related work and methods for charts data extraction are reviewed. We show all components of our model and inference methods in Section~\ref{model}. The computational results are discussed in Section~\ref{results}. 

\section{Literature Review}
\label{literature review}
Automated chart analysis has been studied for many years, and the process of extracting data from charts in documents can be divided into four steps: chart localization and extraction, chart classification, text and element detection, data reconstruction. Our work focuses on the last three steps. 

Chart classification is a specific kind of image classification problems. In 2007, Prasad et al. \yrcite{Prasad2007} have presented a traditional computer vision-based approach to classify charts in five categories. This approach is based on the Histograms of Oriented Gradients and the Scale Invariant Feature Transform descriptors for feature extraction and Support Vector Machine (SVM) for classification. Savva et al. \yrcite{Revision} have built a system called Revision to redesign charts, which includes chart classification, data extraction and visualization. Low-level image features are extracted for chart classification by SVMs.

In recent years, deep learning techniques have made great progress in general image classification \cite{AlexNet, VGG, ResNet}, which can be applied to chart classification. Among these methods, convolutional neural networks based methods are the most widely used, and Siegel et al. \yrcite{FigureSeer} have trained both AlexNet \cite{AlexNet} and ResNet-50 \cite{ResNet} on their annotated datasets including 60,000 charts and 7 categories. VGG16 \cite{VGG} is employed in our system.

A chart includes a set of structural texts and data. In bar charts, texts can be categorized as title, axis-title, axis-tick or legend, and data information is encoded into the height or width of bars. To extract the textual and graphical information from bar charts, one must firstly detect the bounding boxes of texts and bars. Object detection is a common problem in computer vision. With the development of deep learning techniques, there are two main kinds of methods: 1. RCNN \cite{Gir2014RCNN} includes two stages of generating region proposals and subsequent classification; 2. derivations fast-RCNN \cite{Gir2015fastRCNN} and Faster-RCNN \cite{Ren2015FasterRCNN} of RCNN, YOLO \cite{YOLO} and SSD \cite{SSD} use only one stage including both region proposing and classification, which usually perform better on training speed but worse on accuracy of bounding box prediction \cite{ComparisonOD}. It is worth pointing out that Faster-RCNN produces higher accuracy than YOLO and SSD at the expense of a higher training time. There are also some specially designed models \cite{c11,c12} which only focus on text detection. Tian et al. \yrcite{c11} use an anchor box method to predict text bounding boxes. Shi et al. \yrcite{c12} introduce a segment linking method that can handle oriented text detection. 

In terms of chart component detection, there are many works done with traditional computer vision techniques. Zhou et al.  \yrcite{Hough2000} combined Hough transform and boundary tracing to detect bars. Huang et al. \yrcite{Huang2007} have employed rules to detect chart components using edge maps. In \cite{Revision}, bars or pies are detected by their shapes and color information in pixels. By using deep learning techniques, all the texts and chart components can be detected in a model automatically. There are already some works based on deep learning techniques; Cliche et al. \yrcite{Scatteract} have trained three separate object detection models ReInspect \cite{ReInspect} to detect tick marks, tick labels and points in different resolutions, which are finally combined to extract data from a  scatterplot. Poco et al. \yrcite{Poco2017} have employed a CNN to classify each pixel as text or not in a chart and then remove all non-text pixels.

Data reconstruction is followed after chart component detection. Data reconstruction also can be divided into text recognition and numerical data extraction. All detected text bounding boxes are processed by a text recognition model. Tick labels are combined with the coordinate positions to predict the coordinate-value relation. Other graphical components are processed by these coordinate-value relations, which give all the data corresponding to an input chart. There is another challenging problem during this process: matching objects between different categories, which has been studied by a few researchers. In \cite{FigureSeer}, the matching task is formulated as an optimal-path-finding problem with features combined from CNN and pixels. Samira et al. \yrcite{Figureqa} have trained a model with a baseline of Relation Networks (RN) \cite{RN} to build the relationship between objects and answer questions related with it.

In our work, a single deep neural net is built to predict all objects' bounding boxes and classes, to recognize texts of textual objects, and match objects in an chart image, which is very different from all aforementioned works because each prior work on bar and pie charts addresses only a single aspect leading to brittle solutions. We also introduce brand new concepts and approaches such as rotations and angles in conjunction with recurrent neural nets, and supervised angle learning.

\section{Model}
\label{model}
Our model uses Faster-RCNN \cite{Ren2015FasterRCNN} as the object detection part to detect all the elements in a chart, and then uses the idea from RNs \cite{RN} as the object-matching part to match elements between different classes. We use Faster-RCNN over YOLO or SSD because of higher accuracy. The price is higher training time. CRNN \cite{shi2017CRNN} is employed as the text recognition part in our model. All these parts are summarized next.

We build our model on Faster-RCNN because an accurate prediction of bounding box locations is more important in chart component detection compared with general object detection problems.
\subsection{Background}

\subsubsection{Faster-RCNN}

Faster RCNN uses a single convolutional neural network to create feature maps for each predefined regional proposal called also anchor proposal and predicts both a class label and bounding box for each anchor proposal. 
The first part in Faster-RCNN is the Region Proposal Network (RPN), which yields a feature map for each anchor proposal. During the second part, two branches predict a class and a more accurate bounding box location for each anchor proposal. 
In the model, CNN is used to generate a single feature map. Anchor proposals are then selected as follows. A predefined grid of pixels are used as centers of proposals. Each center pixel is associated with a fixed number of bounding boxes of different sizes centered at the pixel. These anchor proposals are identified with the viewing field in the original image to identify the ground truth of the class and the precise bounding box location.

\subsubsection{Relation Networks}

RNs \cite{RN} were proposed by Santoro et al. to solve the relation reasoning problem by a simple neural network module, which is appended after a series of convolutional neural layers. An RN firstly concatenates the object features from the last convolutional layer of two objects and then employs a fully connected layer to predict their true/false relation. The loss function component reads

\begin{equation}
RN(O)  = f_\phi \left ( \frac{1}{N^2} \sum_{i,j}g_\theta(o_i,o_j) \right )
\end{equation}
where \(O \in R^{N\times C} \) is the matrix in which the 
\(i'{th}\) row contains object representation \(o_i\). Here, \(g_\theta\) calculates the relation
between a pair of objects and \(f_\phi\) aggregates these relations and computes the final output of the model.

\subsubsection{CRNN}
CRNN \cite{shi2017CRNN} is a text recognition model, which is based on the popular CNN-RNN pipeline for text recognition. It has gained state-of-art accuracy on several text recognition challenges. This model contains 7 convolutional layers for feature extraction followed by 2 bidirectional LSTM layers for sequence labeling and a transcription layer on top of them.
A sequence is a sequence of characters and the input to an LSTM cell is a fixed width window in the feature map (which keeps sliding from left to right to form a sequence). 

\subsection{Main Model}

The first step in data extraction is to determine the chart types (bar or pie) which consequently involves the corresponding model for each type. The classification model is built based on VGG16.

Faster-RCNN is employed as the backbone of our model,  with its two outputs, a class label, and a bounding box offset, for each candidate object augmented by additional components.
In our bar chart data extraction model, two additional branches are added:
a text recognition branch is added after proposals with a predicted label of text, and an object-matching branch is added for all possible proposal combinations.
Another angle prediction branch is added in the pie chart data extraction model, and the object-matching branch is revised to capture slices.
In bar charts, the RN components try to match bars with a legend component or x-axis label, and the height of a bar with a y-axis value. In pie charts, we attempt to match each slice with a legend component.

In order to handle non-horizontal text such as y-axis labels, a regression layer to predict orientation is added as part of the text recognition branch. 
The feature map is then rotated by the angle and then CRNN is applied.
The object-matching branch is inspired by the idea of RN by concatenating two object features if class predictions are high for the classified two objects. Figure~\ref{fig1} depicts the model.

We next provide details of these components. The first three sections apply to both bar and pie chart data extraction, while the last section explains the enhancements made in pie chart data extraction.

\subsubsection{Object Detection}

Our object detection method is based on Faster-RCNN, which adopts VGG16 \cite{VGG} as the backbone. The top layer of the convolutional feature map (Conv5\_3) is used to generate a manageable number of anchor bounding boxes of all chart components.
Considering that the texts and bars have substantial variety in height/width ratios and can be of arbitrary orientation, and the tick marks are always small, the aspect ratios of anchor boxes are set to (0.1, 0.2, 0.5, 1.0, 2.0, 5.0, 10.0), while the scales are set to (2, 4, 8, 16, 32). Scale of 4 refers to \(4\times 4\) pixels in the feature map.

There is no resizing of the input images because a resizing procedure would lose the resolution for text in the image. Some texts are already very small in the original image and cannot be distinguished even by humans after resizing. The flexible size of the input image affects the size of the resulting convolutional feature maps.
For example, a feature map of size \(m\times n\) generates \(m\times n \times 35\) candidate bounding boxes through the anchor mechanism. Only candidates with an IoU overlap to ground truth higher than a threshold are saved and sorted by their confidence. The top 256 of them are then fed into the regression neural network. 

\subsubsection{Oriented Text Recognition}

The original Faster-RCNN model generates only horizontal bounding boxes, which may contain texts with orientation. Besides the regression layer for localization, a new regression layer for orientation is added under the branch for text recognition, see Figure~\ref{fig1} .

Our text recognition branch is added after the Faster-RCNN model and takes all proposals predicted as text by the classification layer in Faster-RCNN. The text recognition branch detects the orientation of the text in each text bounding box firstly and then rotates the feature map by the detected orientation using an affine transformation, which is similar to the Spatial Transformer Network \cite{Jad2015STN}, except that we use the orientation angle for supervised learning. In summary, the loss function includes the $L_2$ loss $L_{orientation}$ of the angle (the ground truth for angle in training is known).

We apply two-layer bidirectional LSTM to predict the sequence labels of characters, followed by a sequence-to-sequence mapping to words. The conditional probabilities defined in CTC \cite{c19} are adopted as the loss function of the recognition model.

The loss function of the branch for oriented text recognition is
\begin{equation}
L_{text} = \lambda_{o}L_{orientation} + \lambda_CL_{CTC}
\end{equation}

\subsubsection{Object Matching}

There is an object matching branch appended after the feature map (Conv5\_3) generated by Faster-RCNN, which aims to match image patches from different categories.

Our object matching branch is similar to RN but without the summation of feature maps, and the output of RN is normalized to better predict whether the input object pair is related or not
\begin{equation}
OM(o_i,o_j)  = f_\phi \left ( g(o_i,o_j) \right ) \in [0,1]
\end{equation}
where \(g\) is the concatenation operation and \(f_\phi\) refers to a fully connected neural network. Here $o$ is the feature map of an anchor proposal.

The loss function of the object matching branch between types \(U,V\) is formulated as
\begin{equation}
\begin{aligned}
L_{OM} = \sum_{o_i \in U, o_j \in V} & H(P_{o_i})  \cdot H(P_{o_j}) \\ 
&  \cdot KL(y_{o_i,o_j} || softmax(OM({o_i,o_j})))
\end{aligned}
\end{equation}
where \(y_{o_i,o_j}\) is the ground truth, \(KL\) is Kullback--–Leibler divergence and \( H\) is a smooth approximation to the indicator function
\begin{equation}
H(x) = \frac{1}{1+exp(\frac{-k(x-\tau)}{1-\tau})}
\end{equation}
where \( \tau\) is a threshold parameter and \(k\) is another parameter controlling the steepness of the function near \( \tau \). Types \(\{U,V\}\) in bar chart data extraction are \{bar, legend sample\} and in pie chart data extraction are  \{pie, legend\}. We also use the same model to match textual objects with other objects while \(o_i\) in above equations refers to the position of each object. These types of \(\{U,V\}\) cover  \{y-tick label, y-tick mark\} and \{legend sample, legend label\}. We found that this works much better than feature map vectors.

\subsubsection{Pie Chart-Based Enhancements}

Data extraction for pie charts is more challenging than for bar charts since a slice is hard to detect using a rectangular bounding box. Instead of generating each proposal bounding box for each slice, we only predict the location of the whole pie and the angle of each slice's boundary based on the feature map of the pie. Obtaining all the boundary angles gives the information of the proportion of the pie chart.

\begin{figure}[ht]
\vskip 0.2in
\begin{center}
\centerline{\includegraphics[width=0.9\columnwidth]{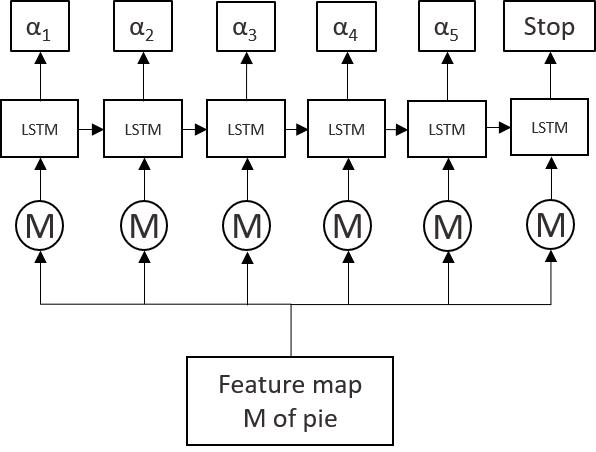}}
\caption{Angle prediction branch for pie chart extraction}
\label{fig2} 
\end{center}
\vskip -0.2in
\end{figure}

Our pie boundary angle prediction model is a two-layer LSTM appended after the feature map of the predicted pie anchor proposal. The predicted pie feature map is fed into the LSTM recurrently until the LSTM outputs a stop signal, while the outputs before the stop signal represent the angles of boundaries in counter-clockwise flow, see Figure~\ref{fig2}. Values $\alpha_1-\alpha_5$ represent the angles in order of  counter-clockwise, with respect to the slice proposals.

The object-matching model between bars and legends in bar chart data extraction relies on the rectangular feature map of each bar or legend, which is not appropriate to match slices and legends for the non-rectangular shapes of slices. 

\begin{figure}[ht]
\vskip 0.2in
\begin{center}
\centerline{\includegraphics[width=0.9\columnwidth]{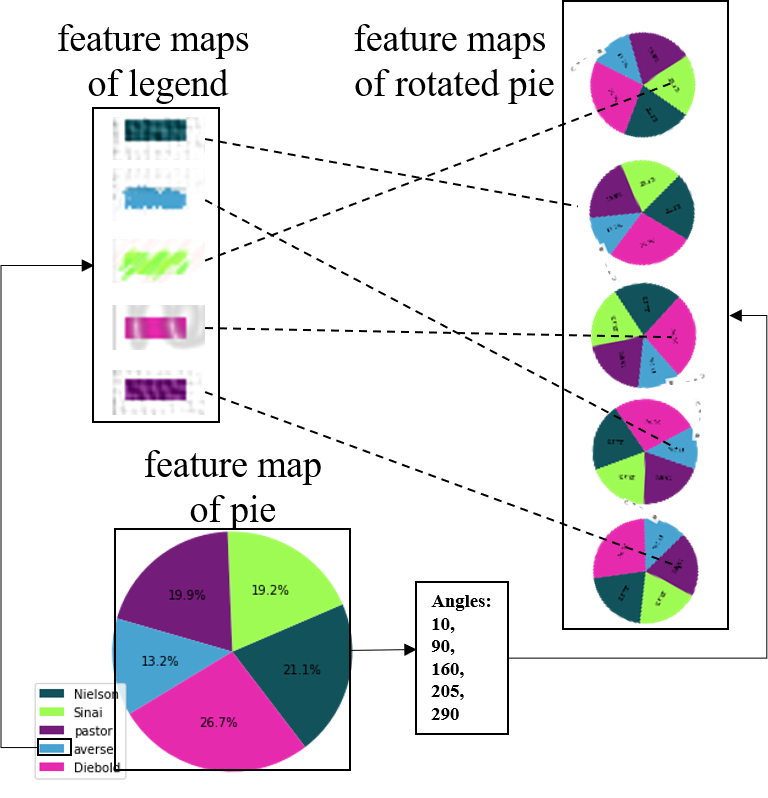}}
\caption{Pie chart object-matching model}
\label{fig3}
\end{center}
\vskip -0.2in
\end{figure}

In the pie chart object-matching model, the feature map for each slice is generated from rotating the feature map of the whole pie by the angle of its boundary, so that the rotated feature map can have its corresponding slice in a specific region. The rotation is done by an affine transformation. We define the horizontal ray from left to right as zero degree and counter-clockwise as the positive direction, so all of the feature maps have their corresponding slice features on the right-center region in the whole feature maps. Figure~\ref{fig3} illustrates the strategy. 

The object matching part is similar to the one for bar charts, which concatenates the feature map of each legend and each rotated feature map of the pie and predicts their relationship. This component learns which region of the feature map is in focus.

\subsubsection{Loss Function}
\label{loss function}
Our loss functions for both bar chart and pie chart data extraction models take the form of a multi-task loss, which are formulated as:
\begin{equation}
L_{bar} = 
L_{det}+\lambda_{text}L_{text}+\lambda_{OM}L_{OM}
\end{equation}
\begin{equation}
L_{pie} = 
L_{det}+\lambda_{text}L_{text}+\lambda_{OM}L_{OM} +\lambda_{ang}L_{ang}
\end{equation}
where $L_{det}$ represents the loss for object detection, which is defined in \cite{Ren2015FasterRCNN}. $L_{text}$ and $L_{OM}$ are the losses for text recognition and object matching defined by (2) and (4), respectively. $L_{ang}$ is the loss for angle prediction in the pie chart data extraction model. The three parameters component $\lambda_{text}$, $\lambda_{OM}$ and $\lambda_{ang}$ balance each loss function component. In our experiment,  $\lambda_{text}$ is set to 0.1, and  $\lambda_{OM}$ and $\lambda_{ang}$ are set to 1.


\subsection{Inference}

The inference approach for bounding box prediction and classification is outlined in \cite{Ren2015FasterRCNN}, and the inference approach for text prediction is specified in \cite{shi2017CRNN}. 

Each chart is first recognized by the chart type classification model to decide which data extraction model to use. 

For a bar chart, each of its anchor proposals with confidence more than 0.8 is generated along with their text prediction and class. Non-maximum suppression is employed to remove redundant proposals and the remaining proposals are fed into the object-matching model to generate the final prediction with inner relationship between proposals.
There is still one last step for bar chart data extraction: linear interpolation is employed to generate the value of a bar from pixel locations to the y-axis value. The predicted values are detected by the top boundary locations of bars.

Inference for a pie chart is slightly different. Since we assume that there is only one pie in a pie chart, we feed the bounding box proposal (pie) with highest confidence after non-maximum suppression into the angle prediction model. The feature map of the pie is then rotated by each predicted angle to match with legend sample feature maps.

\section{Computational Study}
\label{results}
In this section, we first illustrate the training strategies followed by the training and evaluation data generation process for the three deep learning models. 
Then we show the effectiveness of our models by evaluating their performance on general test sets.

\subsection{Training Strategy}
All of the chart type classification model and data extraction models have the backbone of VGG16. We start from the pre-trained ImageNet VGG16 weights as the first training step.

The second training step for chart type classification is to train it on our simulated chart data set described in Section~\ref{data generation}.

In terms of the two chart data extraction models, the second training step focuses on the object detection and classification branches. During this step, parameters $\lambda_{text}, \lambda_{OM}, \lambda_{ang}$ are fixed to zero and the weights of these corresponding branches are fixed to their initial random values. After the training loss of object detection and classification converges, weights in these three branches are released to become trainable and their parameters are set to the values provided in Section~\ref{loss function} and fixed until the end of training. While these fixing steps can be iterated, we observe that a single pass provides a good solution.

\subsection{Training and Evaluation Data Generation}
\label{data generation}
We utilize both simulated and public data sets because of the limited availability of the latter. Matplotlib Python library is used to build the data set for bar and pie chart data extraction models. To have varieties, we introduce randomness in colors for bar or pie, font sizes and types, orientations of texts, sizes of images as well as sizes of bars and pies. All of the titles and x,y-axis labels are assumed to have less than three words from a vocabulary size of 25,000. 
Font type can be any one of 35 commonly used ones. Possible font size ranges for each type of text (titles, axis labels, tick and legend labels) in a chart are set separately, and color choices for bars and pies are arbitrary. 
Note that in the simulation code we output all of the necessary ground truth information, e.g., angles. For bar charts, tick mark and frame styles are also considered with randomness. 
Each bar chart can be a single bar chart or a grouped one with up to five groups. In terms of pie charts, the angle of the first right-center slice boundary is random. The number of slices in our data set is assumed to be less than 10.

The bar chart data extraction model is trained only on our simulated data set of 50,000 images due to the limitation of any public data set with annotations of bounding boxes for bar charts. The annotations for ground truth include classes and bounding boxes for all objects, orientations and texts for textual objects, and object matching relations in charts.

40,000 simulated pie charts are generated by the above strategy. Besides the annotations in bar charts, there is an additional type of labels for boundary angles of slices, which starts from the right-center one and follows the counter-clockwise flow. The training data set also includes 10,000 charts from the Microsoft FigureQA \cite{Figureqa} data set, so the training data set for the pie chart model consists of 50,000 images. FigureQA bar charts are not used in training bar charts since they are too simple and not diversified.

The chart type classification model is trained on our simulated data set of 1,000,000 charts and fined-tuned on a data set consisting of 2,500 charts downloaded from Google Image. The simulated data set is generated by the Microsoft Excel C\# and Matplotlib Python library. The 2,500 chart images are downloaded from Google Image by using keywords ``bar chart," ``pie chart" and labeled based on the corresponding keywords. We use them to fine-tune the model in training.

We use the same stragedy to generate the validation data set of 5,000 bar chart images with annotations for the bar chart model. 4,000 simulated pie charts and 1,000 charts from FigureQA data set are integrated as the validation data set for the pie chart model. The validation data set for chart type classification includes 5,000 simulated charts and 500 downloaded charts.

In order to show our models' effectiveness, we consider the following test data sets: the first one called Simulated is the data set with the same distribution as our training data set and consisting of 3,000 charts in each setting;
the second one for both bar and pie models is 1,000 charts from the FigureQA data set; the third one called Annotated is 10 charts downloaded from Google Image and labeled manually for each data extraction model; the last one called Excel is generated by Microsoft Excel C\# and consists of 1,000 charts for each model.

Regarding Excel C\#, through their API we were not able to figure out how to generate all of the necessary ground truth for training and thus they are not included in training data sets. In test evaluation, we only generate a subset of ground truth labels and use them only for the corresponding metrics.

\subsection{Chart Type Classification}
Our chart type classification model is based on VGG16. The only difference is in the output layer, in which our model has two output categories, ``bar chart" and ``pie chart." The classification model achieves an accuracy of 96.35\% on the test data set of 500 images downloaded from Google.

\subsection{Bar Chart Data Extraction}
The effectiveness of the bar chart data extraction model is first demonstrated by its object detection branch of the 10 categories. As shown in Table~\ref{table1}, the mean average precision for the simulated test data set is 92.6\%, which is higher than 84.5\% for FigureQA and 59.7\% for Annotated. Any of the x-tick lines in the FigureQA data set are not detected since the x-tick lines in the FigureQA charts are much longer than our simulated ones. Legend marks in the Annotated data set are also hard to detect since they have different styles or sizes from our training data set.

\begin{table}[t]
\caption{Average Precision for each object in Simulated (Simul), FigureQA (FigQA) and human annotated (Annot) bar chart data sets}
\label{table1}
\vskip 0.15in
\begin{center}
\begin{small}
\begin{sc}
\begin{tabular}{lrrrr}
\toprule
Object  & Simul & FigQA & Annot \\
\midrule
Title    & 100.0 & 97.3 & 47.5 \\
X-axis label & 99.9 & 94.1 & 38.2 \\
Y-axis label & 90.9 & 93.5 & 75.0 \\
X-tick label & 91.0 & 85.0 & 48.1 \\
Y-tick label & 88.1 & 91.3 & 75.4 \\
X-tick line  & 89.4 &  0.0 & 58.2 \\
Y-tick line  & 87.5 & 87.0 & 76.5 \\
legend label & 90.8 & 100.0 & 75.1 \\
legend mark  & 90.2 & 100.0 & 24.3 \\
bar          & 97.9 & 96.8 & 80.2 \\
Mean         & 92.6 & 84.5 & 59.7 \\
\bottomrule
\end{tabular}
\end{sc}
\end{small}
\end{center}
\vskip -0.1in
\end{table}

To further evaluate the performance of the bar chart data extraction model, we propose the following evaluation metric. We use a 4-tuple to represent the prediction of each bar in a bar chart: (x-tick label, value, lower y-tick label, upper y-tick label). In the 4-tuple, x-tick label is matched to the bar by the object matching branch (x-tick label can be either below the bar or in the legend), value is predicted using the top boundary location of the bar. The introduction of the lower and upper y-tick labels makes the evaluation more reliable since the predicted value can slightly differ from the ground truth. The lower y-tick label is the y-tick label immediately below the actual value of the bar (and similarly defined upper y-tick label). Besides the 4-tuple, each bar chart has its prediction of title, and x- and y-axis labels. All the prediction results are summarized in Table~\ref{table2}.

In Table~\ref{table2}, the accuracy of the entire ``ALL" charts, titles, and x-, y-axis labels in the first 4 rows represent the percentage of correctly predicted chart instances in each test data set. In all elements consisting of text we count true positive only if all words match correctly. We define true positive charts ("ALL") as predicted with correct title, x-, y-axis labels and all 4-tuples with less than 1\% error in value.
The following 11 rows are based on each tuple prediction. For example, the true positive instance for ``tuple 10\% err" means the tuple is predicted with a less than 10\% error in value and the remaining 3 textual elements are completely correct. The error is defined as:
\begin{equation}
Error = \frac{|Value_{pred}-Value_{GT}|}{|Value_{GT}|}
\end{equation}
where $Value_{GT}$ is the ground truth value of each bar.

\begin{table}
\caption{Accuracy of bar chart prediction in Simulated (Simul), FigureQA (FigQA), annotated (Annot) and Excel bar chart data sets}
\label{table2}
\vskip 0.15in
\begin{center}
\begin{small}
\begin{sc}
\begin{tabular}{lrrrr}
\toprule
Object  & Simul & FigQA & Annot & Excel\\
\midrule
ALL  & 79.4 & 69.2 & 30.0 & 68.3\\
Title & 95.6 & 94.8 & 40.0 & 91.0 \\

X-axis label & 94.9 &93.7&	40.0&	89.2\\
Y-axis label & 88.2	&93.0&	50.0&	83.1 \\
Tuple 1\% err & 81.0&	72.3&	28.4&	77.3\\
Tuple 5\% err & 83.4&	76.2&	32.8&	79.2\\
Tuple 10\% err & 83.9&	78.4&	34.3&	69.8\\
Tuple 25\% err & 85.6&	81.3&	38.8&	83.4\\
X-tick label & 87.3	&83.5&	43.3&	87.5\\
Lower value & 87.8&	85.1&	56.0&	83.2\\
Upper value & 88.3	&85.4&	58.2&	84.9\\
Value 1\% err & 82.4&	77.9&	41.0&	80.2\\
Value 5\% err & 86.0&	81.2&	46.3&	82.9\\
Value 10\% err & 87.7&	83.4&	50.0&	83.3\\
Value 25\% err & 91.3&	89.1&	61.2&	88.0\\
\bottomrule
\end{tabular}
\end{sc}
\end{small}
\end{center}
\vskip -0.1in
\end{table}

The results show the model works well on Simulated, FigureQA and Excel bar chart data sets but not as good on the Annotated data set. 
\begin{figure}[t]
\centering
  \begin{subfigure}{0.49\textwidth}
    \centering
    \includegraphics[width=\textwidth]{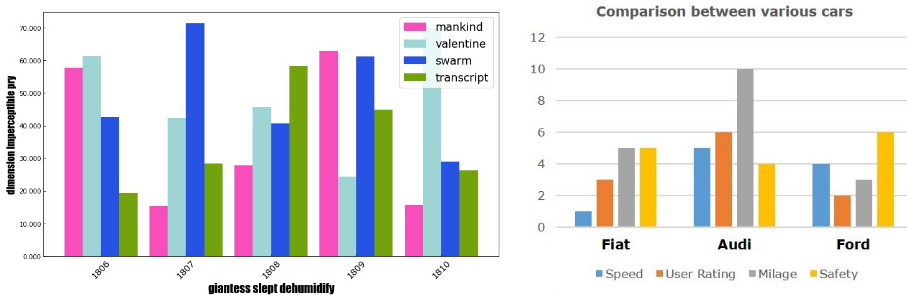}
    \caption{Bar chart samples with ``good" predictions, left from Simulated, right from Annotated}
    \label{fig:4a}
  \end{subfigure}
  \begin{subfigure}{0.49\textwidth}
    \centering
    \includegraphics[width=\textwidth]{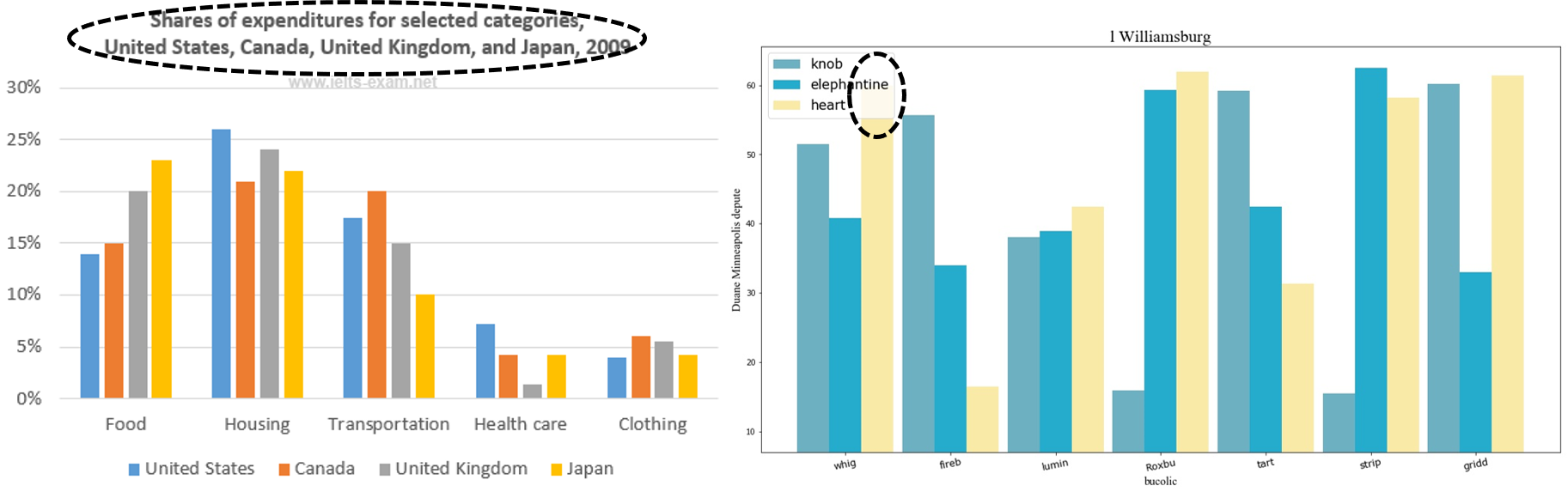}
    \caption{Bar chart samples with ``OK" predictions, left from Annotated, right from Simulated}
    \label{fig:4b}
  \end{subfigure}  \begin{subfigure}{0.49\textwidth}
    \centering
    \includegraphics[width=\textwidth]{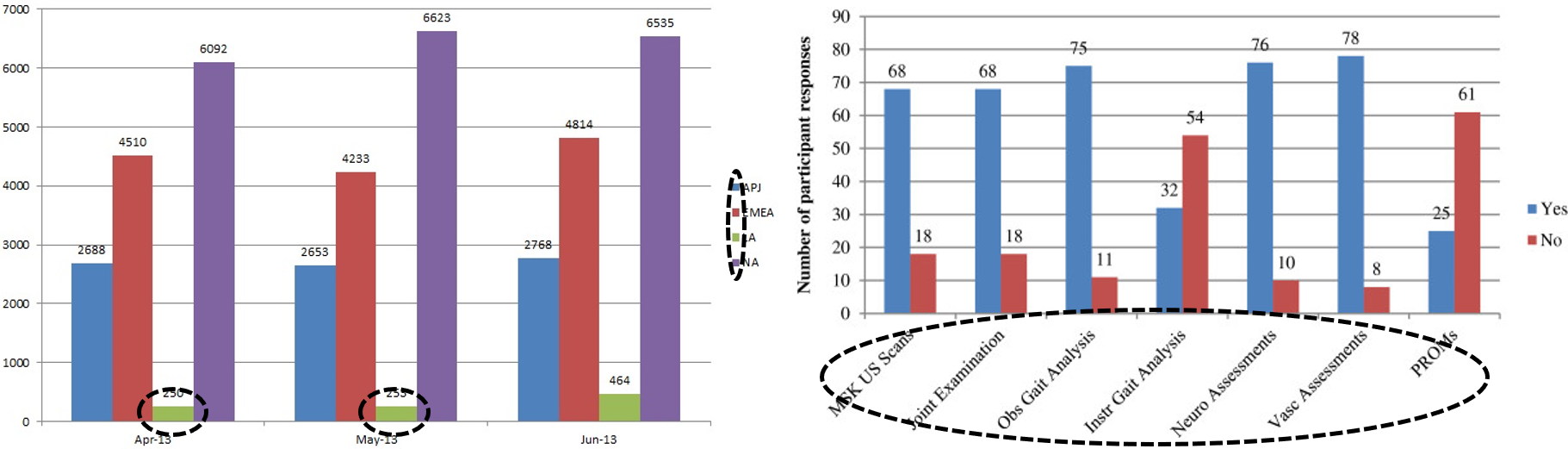}
    \caption{Bar chart samples with ``bad" predictions, both from Annotated}
    \label{fig:4c}
  \end{subfigure}
  \caption{Bar chart samples with ``good," ``OK" and ``bad" predictions; dash circles indicate problematic regions}
  \label{fig:4}
\end{figure}

We also plot six chart samples with ``good," ``OK" or ``bad" predictions. ``Good" samples are from true positives, while ``OK" samples predict some parts wrongly. ``Bad" samples miss some important objects like bars or x-tick labels. The model on the sample on the left in 
Figure~\ref{fig:4b} cannot detect the entire 2-line title of more than 10 words since we do not have such training samples, and the bar on the right chart cannot be detected correctly since the legend region covers it. The left ``bad" sample in 
Figure~\ref{fig:4c} has two very low bars and very small legend marks which are not detected by the model. The right one has several long x-tick labels that are not detected correctly. These wrong predictions are mainly caused by lack of variance in the training data set or the convolution operations not working well on small objects. A more accurate multi-line and multi-word text detection and recognition model would improve the model. This can be handled by augmenting the training data set. It is unclear how to improve the performance on small objects.

\subsection{Pie Chart Data Extraction}

Following similar evaluation metrics for bar charts, we first evaluate the average precision of 4 categories and their mean value for pie charts in Table~\ref{table_pie_map}.

Compared with bar charts, objects in pie charts are much easier to detect since there are fewer categories and the sizes of objects are larger.

Different from bar charts, we define a 2-tuple to represent the prediction of each slice in a pie chart, (legend, percentage). The accuracy of our prediction results is shown in Table~\ref{table_pie_res}. The error for pie charts is defined as:
\begin{equation}
Error = \frac{|Percentage_{pred}-Percentage_{GT}|}{Percentage_{GT}}.
\end{equation}
The results show a great success on the FigureQA data set since their pies have less variance and are included in our training data set. The whole accuracy for the Annotated data set is much worse than the other two data sets because the accurate prediction of percentage of slices is hard, although it gets a high accuracy of 77.8\% for percentage prediction in the Annotated data set with less than 5\% errors.

\begin{figure}
\centering
  \begin{subfigure}{0.49\textwidth}
    \centering
    \includegraphics[width=\textwidth]{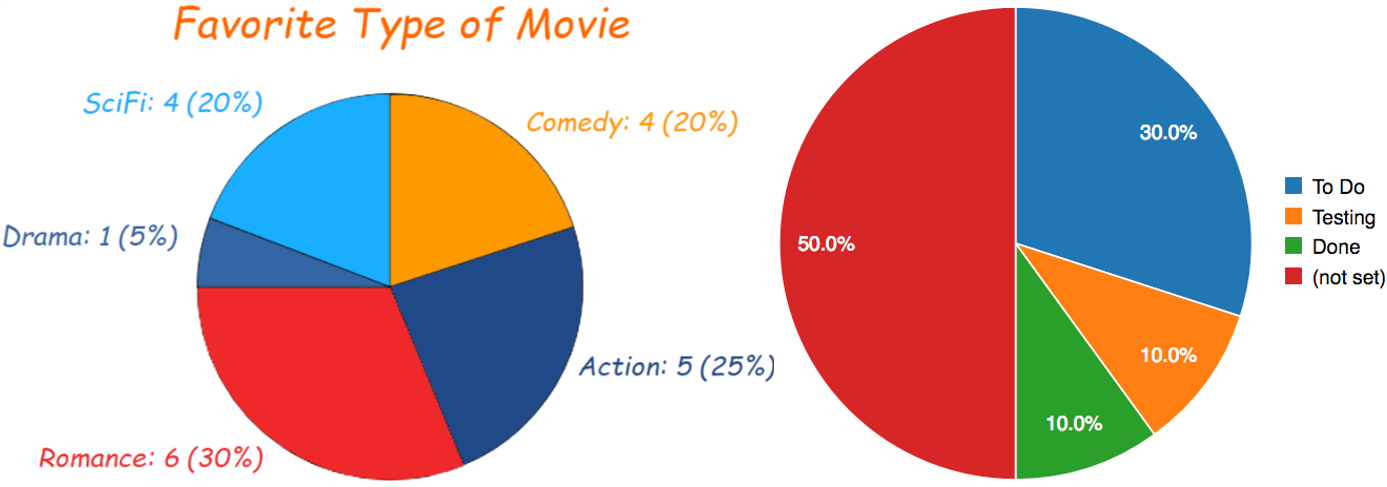}
    \caption{Pie chart samples with ``good" predictions, from Annotated}
    \label{fig:5a}
  \end{subfigure}
  \begin{subfigure}{0.49\textwidth}
    \centering
    \includegraphics[width=\textwidth]{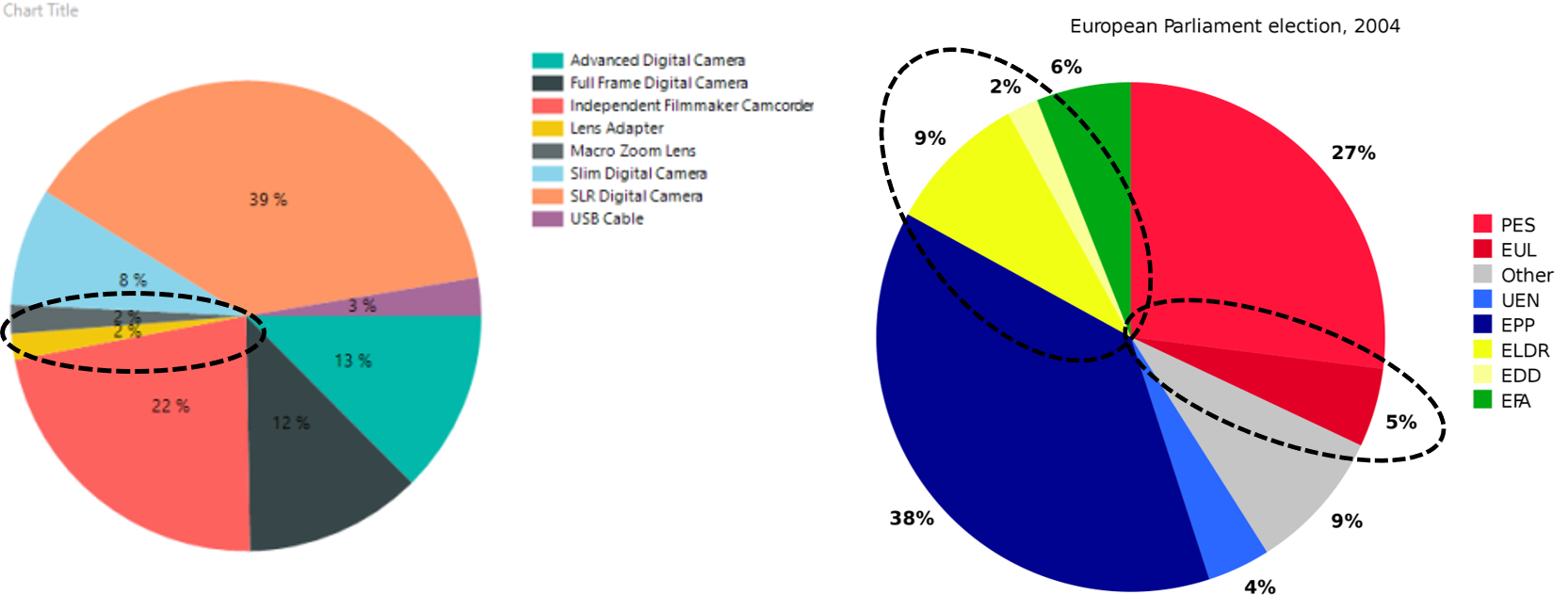}
    \caption{Pie chart samples with ``bad" predictions, from Annotated}
    \label{fig:5b}
  \end{subfigure}  
  \caption{Pie chart samples with ``good" and ``bad" predictions; dash circles indicate challenging regions}
  \label{fig:5}
\end{figure}

Figure~\ref{fig:5} shows 4 pie charts from the Annotated data set which are divided into ``good" and ``bad" samples based on the quality of the predictions. The model works well when the slices have high contrast colors and they are not too narrow. When there are slices with percentage of less than 2\% (left sample in 
Figure~\ref{fig:5b}) or of similar color on its neighboring slice (right sample in 
Figure~\ref{fig:5b}), the model has difficulties handling it.

\begin{table}
\caption{Average Precision for each object in Simulated (Simul), FigureQA (FigQA) and human annotated (Annot) pie chart data sets}
\label{table_pie_map}
\vskip 0.15in
\begin{center}
\begin{small}
\begin{sc}
\begin{tabular}{lrrrr}
\toprule
Object  & Simul & FigQA & Annot \\
\midrule
Title    & 100.0 & 100.0 & 90.0 \\
Pie & 100.0 & 100.0 & 100.0 \\
legend label & 98.9 & 100.0 & 83.3 \\
legend mark  & 95.3 & 100.0 & 66.7 \\
Mean         & 98.6 & 100.0 & 80.4\\
\bottomrule
\end{tabular}
\end{sc}
\end{small}
\end{center}
\vskip -0.1in
\end{table}

\begin{table}
\caption{Accuracy of pie chart prediction in Simulated (Simul), FigureQA (FigQA), annotated (Annot) and Excel pie chart data sets}
\label{table_pie_res}
\vskip 0.15in
\begin{center}
\begin{small}
\begin{sc}
\begin{tabular}{lrrrr}
\toprule
Object  & Simul & FigQA & Annot  & Excel\\
\midrule
ALL  & 88.0 &	98.5& 20.0  & 68.6\\
Title & 96.4& 100.0& 60.0 & 92.5\\
Tuple 1\% err & 90.0&	98.5&	28.9 & 71.2\\
Tuple 5\% err & 90.5&	99.0& 57.8 & 75.3\\
Tuple 10\% err & 90.8&	99.1&	60.0 & 78.0
\\
Tuple 25\% err & 91.8&	99.7&	64.4 & 80.5
\\
Legend & 92.5&	100.0&	68.9 & 87.2
\\
Percent 1\% err & 97.4&	98.5&	33.3 & 83.4
\\
Percent 5\% err & 97.8&	99.0&	77.8 & 88.9
\\
Percent 10\% err & 98.0&	99.1&	80.0 & 91.2
\\
Percent 25\% err & 98.9&	99.7&	86.7 & 93.0
\\
\bottomrule
\end{tabular}
\end{sc}
\end{small}
\end{center}
\vskip -0.1in
\end{table}

\section{Conclusion}
In this paper, a system of neural networks is established to classify chart types and extract data from bar and pie charts. The data extraction model is a single deep neural net based on Faster-RCNN object detection model. We have extended it with text recognition and object matching branches. We have also proposed a percentage prediction model for pie charts as another contribution. The model has been trained on a simulated data set and performs successfully on 79.4\% of the simulated bar charts and 88.0\% of the simulated pie charts. The performance on the images downloaded from Internet is worse than on the simulated data set or another generated data set since it includes more variantions not seen in training. Augmenting the training data set by means of a more comprehensive simulation (or Microsoft offering addition API functionality in Excel C\#) is an easy way to substantially improve the performance. It is more challenging to find a way to cope with small objects.


\section*{Acknowledgements}

This work was supported by Intel Corporation,
Semiconductor Research Corporation (SRC).



\begin{thebibliography}{24}
\providecommand{\natexlab}[1]{#1}
\providecommand{\url}[1]{\texttt{#1}}
\expandafter\ifx\csname urlstyle\endcsname\relax
  \providecommand{\doi}[1]{doi: #1}\else
  \providecommand{\doi}{doi: \begingroup \urlstyle{rm}\Url}\fi

\bibitem[Cliche \& Yee(2017)Cliche and Yee]{Scatteract}
Cliche, M., Rosenberg D. Madeka~D. and Yee, C.
\newblock Scatteract: Automated extraction of data from scatter plots.
\newblock In \emph{Joint European Conference on Machine Learning and Knowledge
  Discovery in Databases}, pp.\  135--150, 2017.

\bibitem[Girshick \& Malik(2014)Girshick and Malik]{Gir2014RCNN}
Girshick, R., Donahue J. Darrell~T. and Malik, J.
\newblock Rich feature hierarchies for accurate object detection and semantic
  segmentation.
\newblock In \emph{CVPR}, pp.\  580--587, 2014.

\bibitem[Girshick(2015)]{Gir2015fastRCNN}
Girshick, R.
\newblock Fast R-CNN.
\newblock In \emph{CVPR}, pp.\  1440--1448, 2015.

\bibitem[Graves \& Schmidhuber(2006)Graves and Schmidhuber]{c19}
Graves, A., FernÃ¡ndez S. Gomez~F. and Schmidhuber, J.
\newblock Connectionist temporal classification: labelling unsegmented sequence
  data with recurrent neural networks.
\newblock In \emph{ICML}, pp.\  369--376, 2006.

\bibitem[He \& Sun(2016)He and Sun]{ResNet}
He, K., Zhang X. Ren~S. and Sun, J.
\newblock Deep residual learning for image recognition.
\newblock In \emph{CVPR}, pp.\  770--778, 2016.

\bibitem[Huang \& Murphy(2017)Huang and Murphy]{ComparisonOD}
Huang, J., Rathod V. Sun C. Zhu M. Korattikara A. Fathi A. Fischer I. Wojna Z.
  Song Y. Guadarrama~S. and Murphy, K.
\newblock Speed/accuracy trade-offs for modern convolutional object detectors.
\newblock In \emph{CVPR}, pp.\  7310--7319, 2017.

\bibitem[Huang \& Tan(2007)Huang and Tan]{Huang2007}
Huang, W. and Tan, C.L.
\newblock A system for understanding imaged infographics and its applications.
\newblock In \emph{ACM Symposium on Document Engineering}, pp.\  9--18, 2007.

\bibitem[Jaderberg \& Zisserman(2015)Jaderberg and Zisserman]{Jad2015STN}
Jaderberg, M., Simonyan~K. and Zisserman, A.
\newblock Spatial transformer networks.
\newblock In \emph{NIPS}, pp.\  2017--2025, 2015.

\bibitem[Kahou \& Bengio(2017)Kahou and Bengio]{Figureqa}
Kahou, S.E., Atkinson A. Michalski V. KÃ¡dÃ¡r Ã. Trischler~A. and Bengio, Y.
\newblock Figureqa: An annotated figure dataset for visual reasoning.
\newblock \emph{arXiv preprint arXiv:1710.07300}, 2017.

\bibitem[Krizhevsky et~al.(2012)Krizhevsky, Sutskever, and Hinton]{AlexNet}
Krizhevsky, A., Sutskever, I., and Hinton, G.E.
\newblock Imagenet classification with deep convolutional neural networks.
\newblock In \emph{NIPS}, pp.\  1106--1114, 2012.

\bibitem[Liu \& Berg(2016)Liu and Berg]{SSD}
Liu, W., Anguelov D. Erhan D. Szegedy C. Reed S. Fu~C.Y. and Berg, A.C.
\newblock SSD: Single shot multibox detector.
\newblock In \emph{ECCV}, pp.\  21--37, 2016.

\bibitem[Poco \& Heer(2017)Poco and Heer]{Poco2017}
Poco, J. and Heer, J.
\newblock Reverse-â€engineering visualizations: Recovering visual encodings
  from chart images.
\newblock In \emph{Computer Graphics Forum}, pp.\  353--363, 2017.

\bibitem[Prasad et~al.(2007)Prasad, Siddiquie, Golbeck, and Davis]{Prasad2007}
Prasad, V. S.~N., Siddiquie, B., Golbeck, J., and Davis, L.~S.
\newblock Classifying computer generated charts.
\newblock In \emph{Content-Based Multimedia Indexing Workshop}, pp.\  85--92,
  2007.

\bibitem[Redmon \& Farhadi(2016)Redmon and Farhadi]{YOLO}
Redmon, J., Divvala S. Girshick~R. and Farhadi, A.
\newblock You only look once: Unified, real-time object detection.
\newblock In \emph{CVPR}, pp.\  779--788, 2016.

\bibitem[Ren \& Sun(2015)Ren and Sun]{Ren2015FasterRCNN}
Ren, S., He K. Girshick~R. and Sun, J.
\newblock Faster R-CNN: Towards real-time object detection with
  region proposal networks.
\newblock In \emph{NIPS}, pp.\  91--99, 2015.

\bibitem[Santoro \& Lillicrap(2017)Santoro and Lillicrap]{RN}
Santoro, A., Raposo D. Barrett D.G. Malinowski M. Pascanu R. Battaglia~P. and
  Lillicrap, T.
\newblock A simple neural network module for relational reasoning.
\newblock In \emph{NIPS}, pp.\  4967--4976, 2017.

\bibitem[Savva et~al.(2011)Savva, Kong, Chhajta, Fei-Fei, Agrawala, and
  Heer]{Revision}
Savva, M., Kong, N., Chhajta, A., Fei-Fei, L., Agrawala, M., and Heer, J.
\newblock Revision: Automated classification, analysis and redesign of chart
  images.
\newblock In \emph{Proceedings of the 24th Annual ACM Symposium on User
  Interface Software and Technology}, pp.\  393--402, 2011.

\bibitem[Shi \& Belongie(2017)Shi and Belongie]{c12}
Shi, B., Bai~X. and Belongie, S.
\newblock Detecting oriented text in natural images by linking segments.
\newblock In \emph{CVPR}, pp.\  3482--3490, 2017.

\bibitem[Shi \& Yao(2017)Shi and Yao]{shi2017CRNN}
Shi, B., Bai~X. and Yao, C.
\newblock An end-to-end trainable neural network for image-based sequence
  recognition and its application to scene text recognition.
\newblock In \emph{PAMI}, pp.\  2298--2304, 2017.

\bibitem[Siegel \& Farhadi(2016)Siegel and Farhadi]{FigureSeer}
Siegel, N., Horvitz Z. Levin R. Divvala~S. and Farhadi, A.
\newblock Figureseer: Parsing result-figures in research papers.
\newblock In \emph{ECCV}, pp.\  664--680, 2016.

\bibitem[Simonyan \& Zisserman(2012)Simonyan and Zisserman]{VGG}
Simonyan, K. and Zisserman, A.
\newblock Very deep convolutional networks for large-scale image recognition.
\newblock In \emph{ICLR}, 2012.

\bibitem[Stewart \& Ng(2016)Stewart and Ng]{ReInspect}
Stewart, R., Andriluka~M. and Ng, A.Y.
\newblock End-to-end people detection in crowded scenes.
\newblock In \emph{CVPR}, pp.\  2325--2333, 2016.

\bibitem[Tian \& Qiao(2016)Tian and Qiao]{c11}
Tian, Z., Huang W. He T. He~P. and Qiao, Y.
\newblock Detecting text in natural image with connectionist text proposal
  network.
\newblock In \emph{ECCV}, pp.\  56--72, 2016.

\bibitem[Zhou \& Tan(2000)Zhou and Tan]{Hough2000}
Zhou, Y. and Tan, C.L.
\newblock Hough-based model for recognizing bar charts in document images.
\newblock In \emph{SPIE}, pp.\  333--341, 2000.

\end{thebibliography}

\appendix


\end{document}